\documentclass[conference]{IEEEtran}

\PassOptionsToPackage{bookmarks=false}{hyperref}

\usepackage{cite}
\usepackage{hyperref}
\hypersetup{hidelinks}

\usepackage{orcidlink}
\usepackage{amsmath,amssymb,amsfonts}
\usepackage{algorithmic}
\usepackage{algorithm}
\usepackage{array}
\usepackage[caption=false,font=normalsize,labelfont=sf,textfont=sf]{subfig}
\usepackage{textcomp}
\usepackage{stfloats}
\usepackage{multirow}
\usepackage{makecell}
\usepackage{url}
\usepackage{verbatim}
\usepackage{graphicx}
\usepackage{balance}
\usepackage{xcolor}
\usepackage{booktabs}
\usepackage{balance}

\def\BibTeX{{\rm B\kern-.05em{\sc i\kern-.025em b}\kern-.08em
    T\kern-.1667em\lower.7ex\hbox{E}\kern-.125emX}}

\DeclareMathOperator*{\mean}{mean}

\DeclareRobustCommand{\etal}{\textit{et al.}}

\begin{document}





\title{Multi-vessel Interaction-Aware Trajectory Prediction and Collision Risk Assessment}


\author{
\centering
\IEEEauthorblockN{
Md Mahbub Alam\IEEEauthorrefmark{1},
Jose F. Rodrigues-Jr\IEEEauthorrefmark{2},
Gabriel Spadon\IEEEauthorrefmark{1}
}

\IEEEauthorblockA{\IEEEauthorrefmark{1}\textit{Faculty of Computer Science}, \textit{Dalhousie University}, \textit{Halifax}, \textit{Canada}}

\IEEEauthorblockA{\IEEEauthorrefmark{2}\textit{Computer Science Department}, \textit{University of São Paulo}, \textit{São Carlos}, \textit{Brazil} \\
mahbub.alam@dal.ca, junio@icmc.usp.br, spadon@dal.ca}
}

\maketitle

\begin{abstract}
Accurate vessel trajectory prediction is essential for enhancing situational awareness and preventing collisions. Still, existing data-driven models are constrained mainly to single-vessel forecasting, overlooking vessel interactions, navigation rules, and explicit collision risk assessment. We present a transformer-based framework for multi-vessel trajectory prediction with integrated collision risk analysis. For a given target vessel, the framework identifies nearby vessels. It jointly predicts their future trajectories through parallel streams encoding kinematic and derived physical features, causal convolutions for temporal locality, spatial transformations for positional encoding, and hybrid positional embeddings that capture both local motion patterns and long-range dependencies. Evaluated on large-scale real-world AIS data using joint multi-vessel metrics, the model demonstrates superior forecasting capabilities beyond traditional single-vessel displacement errors. By simulating interactions among predicted trajectories, the framework further quantifies potential collision risks, offering actionable insights to strengthen maritime safety and decision support.
\end{abstract}


\section{Introduction}

Maritime shipping is critical not only for global trade and economy but also for various socio-economic activities, including fishing, passenger transportation, and recreational sailing~\cite{schnurr2019marine}. To enhance navigational safety, the International Maritime Organization (IMO) mandated the use of the Automatic Identification System (AIS) in 2003, with satellite AIS integration in 2008, further expanding monitoring coverage~\cite{Ed2004_consolidated, fournier2018past}. Consequently, the widespread adoption of AIS generates a vast volume of vessel movement data, which has spurred research to address maritime challenges. As maritime traffic has surged in recent years, leading to an increase in vessel collisions often due to human error~\cite{antao2023quantitative, wu2022review}, Knowledge of the future movements of vessels surrounding a navigator is critical for safe navigation. Accurate and timely prediction of vessel movements is essential not only for improving navigators’ situational awareness and reducing collision risks, but also for enabling applications such as optimal route planning, anomaly detection, and large-scale traffic management~\cite{tu2017exploiting, xiao2019traffic}.

Over the past few years, research on vessel trajectory prediction has shifted from physics only model to data-driven models, primarily by leveraging machine learning techniques~\cite{zhang2022vessel}. 
More recently, deep learning has achieved significant improvements in both prediction accuracy and computational efficiency~\cite{li2023ais}. However, most of these efforts have focused on predicting the trajectory of a single target vessel, often overlooking the complex interactions among multiple vessels and governing navigational rules in shared waterways, where mutual awareness and coordination are necessary for safety. Although some recent approaches have incorporated graph-based architectures to model inter-vessel dependencies~\cite{yang2024graph, jiang2024stmgf}, they frequently exhibit several limitations, including challenges in effectively generalizing to complex and large-scale interaction patterns, neglected kinematic features grounded in vessel dynamics, and reliance on trajectory-centric evaluation metrics that do not adequately capture the essence of joint interaction modeling~\cite{weng2023joint}. Furthermore, there is a lack of established guidelines for the effective use of joint predictions in real-world potential collision risk analyses. 

We propose a general framework for modeling multi-agent interactions that uses dynamic kinematic features for trajectory prediction and collision risk assessment. At its core is a Transformer-based model that combines navigational and physics-derived features through parallel processing streams. The main navigational features are handled with causal convolutions, which capture local temporal patterns, and spatial projections, which encode positional coordinates. A hybrid positional encoding is then applied to improve adaptability and generalization. The physics-derived features are passed through a non-linear dense layer, adding knowledge of vessel dynamics and physical constraints. Finally, the two feature streams are fused, enabling the model to capture both short-term motion and long-range dependencies for accurate multi-step trajectory forecasting.


The trained Transformer-based model jointly predicts the target vessel and its neighbors in parallel using DASK~\cite{rocklin2015dask}. For collision-risk assessment, the framework adapts the Closest Point of Approach (CPA)~\cite{zheng2022comprehensive} to operate on joint trajectory predictions. The main contributions are summarized as follows:

\begin{itemize}
    \item We propose a unified framework that models multi-vessel interactions while integrating both kinematic and physics-derived features, enabling realistic trajectory forecasting.
    
    \item We introduce a Transformer-based architecture with hybrid positional encoding that significantly improves prediction accuracy, particularly for long horizons.
    
    \item We validate the framework through a comprehensive evaluation using on joint displacement error metrics. \\
    
    \item We demonstrate practical collision risk assessment by analyzing distance to CPA (DCPA) and time to CPA (TCPA) based on predicted trajectories, showing the framework's potential to support real-time navigational safety.
\end{itemize}

To present our contributions, the remainder of the paper is organized as follows: Section~\ref{sec:related-work} provides a detailed overview of recent works on vessel trajectory prediction; Section~\ref{sec:def-and-notation} explains preliminaries and notations related to the trajectory prediction problem; Section~\ref{sec:framework} presents the proposed multi-vessel collision avoidance framework, including the transformer-based prediction model; Section~\ref{sec:exp-evaluation} provides a detailed experimental evaluation and analyzes both prediction performance and collision risk; and, finally, Section~\ref{sec:conclusion} concludes the paper with a summary of results and future research directions.

\section{Related Work}
\label{sec:related-work}

Methods for vessel trajectory prediction can be broadly categorized into two main groups: traditional model-based and data-driven approaches. Traditional methods are primarily based on mathematical equations (e.g., CVMs and NCVMs), also known as physical models~\cite{tu2017exploiting}. Data-driven methods use historical data to learn mobility patterns and are further divided into classical machine learning (ML) techniques (e.g., Random Forest, XGBoost, MLP) and deep learning (DL) models (e.g., LSTM, GRU, CNN, TCN, Transformers)~\cite{zhang2022vessel}. While early research predominantly relied on physical models, this field shifted toward ML-based data-driven methods to improve prediction accuracy. In recent years, we have seen significant accuracy improvements with DL-based methods~\cite{li2023ais}. This section summarizes recently proposed DL-based approaches to address the research gap. The evolution of these methods began with RNN-based architectures, such as LSTMs and GRUs~\cite{chen2023tdv, slaughter2025vessel}, expanded to include CNNs~\cite{shin2024deep}, and more recently incorporated advanced architectures, such as Temporal Convolutional Networks (TCNs)~\cite{lin2023ship}, Graph Neural Networks (GNNs)~\cite{yang2024graph}, and Transformers~\cite{nguyen2024transformer}.

Early approaches primarily focused on modeling spatial and temporal dependencies inherent in trajectory data. Accordingly, Chondrodima \etal~(2023)~\cite{chondrodima2023efficient} proposed an LSTM-based framework in which a new trajectory augmentation technique was integrated to improve prediction accuracy. As the Seq2Seq model can predict longer sequences efficiently, Forti \etal~(2020)~\cite{forti2020prediction} proposed an LSTM encoder-decoder architecture, whereas Capobianco \etal~(2021)~\cite{capobianco2021deep} used bidirectional LSTM to encode past vessel trajectories and unidirectional LSTM to decode future positions. The model integrates various aggregation mechanisms, including max pooling, average pooling, and attention, to enhance predictive accuracy. To capture both spatial and temporal dependencies efficiently, some approaches combine CNNs with RNN-based architectures. For instance, Dijt \etal~(2020)~\cite{dijt2020trajectory} proposed a multitask Seq2Seq model implemented as a recurrent CNN, where LSTM was used as a recurrent cell, whereas Wu \etal~(2023)~\cite{wu2023ship} utilized Seq2Seq ConvLSTM, which incorporated preprocessing techniques, such as kinematics-based anomaly detection, Douglas–Peucker compression, and HDBSCAN clustering, and represented trajectories as spatial grids to enhance the model's learning capability. 

As prediction errors accumulate over time in RNN-based models, Chen \etal~(2022)~\cite{chen2022multi} proposed a multi-memory LSTM (MM-LSTM) model based on a multi-decoder recurrent network. MM-LSTM introduces global memory states to enhance temporal feature extraction, whereas the multi-decoder structure mitigates long-term error accumulation. Further exploring advanced architectures for long-term prediction, Spadon \etal~(2024)~\cite{spadon2024multi} presented a framework that combines a probabilistic module for predicting routes via a hexagonal grid using spatially engineered probabilistic features, with a deep autoencoder based on CNNs, Bi-LSTMs, and attention for trajectory reconstruction. Beyond recurrent networks, Temporal Convolutional Networks (TCNs) have also shown promise; for instance, Lin \etal~(2023)~\cite{lin2023ship} propose TTAG (Tiered-TCN-Attention-GRU), a hybrid DL model that incorporates a Tiered-TCN (TTCN), with parallel dilated causal convolutions to capture temporal dependencies effectively.

More recently, researchers have started using Transformers to model long-range dependencies. Jiang \etal~(2023)~\cite{jiang2023trfm} proposed TRFM-LS, a hybrid Transformer-LSTM model aimed at improving long-term trajectory prediction accuracy. TRFM-LS leverages the transformer’s multi-head attention to address limitations in capturing distant sequences and uses LSTM to enhance temporal feature embeddings. This approach also incorporates a time-window panning and smoothing filtering method to pre-process noisy and irregular AIS trajectories, thereby ensuring data continuity and temporal ordering. Building on the capabilities of Transformers, Xue \etal~(2024)~\cite{xue2024g} propose G-Trans, a hierarchical approach integrating GRU and Transformer for accurate and computationally efficient vessel trajectory prediction. This approach incorporates preprocessing, including trajectory smoothing and latent-space clustering of motion features, to enhance the understanding of dynamic characteristics. In a related development, Nguyen and Fablet~(2024)~\cite{nguyen2024transformer} introduced TrAISformer, a Transformer-based model for long-term vessel trajectory prediction. It addresses data heterogeneity and multimodality through a sparse high-dimensional \textit{four-hot} trajectory representation, and reframes prediction as a classification task using a custom cross-entropy loss. Beyond Transformers, more recent advances in sequence modeling have led to novel architectures; for example, Suo \etal~(2024)~\cite{suo2024mamba} proposed a deep-learning framework based on Mamba that uses a selective state-space model to process long sequential data efficiently.

However, these approaches focus on predicting the trajectory of a single target vessel, overlooking the crucial role of vessel interactions and navigation rules on shared waterways for situational awareness and traffic management. Therefore, the simultaneous prediction of both the target vessel and its neighbors is critical for enhancing navigational safety. Consequently, researchers are increasingly using Graph Neural Networks (GNNs) to model these interactions. Yang \etal~(2024)~\cite{yang2024graph} proposed GMLTP, a graph-driven framework for long-term multi-vessel trajectory prediction. It integrates a Multi-Graph Spatial Convolution layer to capture complex spatial interactions among vessels, and a ProbSparse Self-Attention Transformer for efficient temporal modeling. Another notable GNN-based contribution is by Zeng \etal~(2024)~\cite{zeng2024trajectories}, who propose ST-GRUA, a spatio-temporal GCN with GRU and Self-Attention, for predicting trajectories in multi-ship encounters. The model leverages GCN for complex spatial interactions and GRU with Attention for temporal dynamics on graph representations derived from AIS data. Experiments showed that ST-GRUA achieves superior accuracy for trajectory, speed, and heading predictions in complex scenarios. 

Further advancing spatio-temporal modeling in multi-vessel scenarios, Wang \etal~(2024)~\cite{wang2024vessel} proposed DAA-SGCN, a deep attention-aware spatio-temporal GCN for predicting trajectories in complex maritime scenarios. The model integrates motion feature encoding via LSTM, spatial interaction modeling through ST-GCN, and temporal dependence extraction using a residual TCN. A novel attention-based weighted adjacency matrix dynamically captures vessel interactions, whereas BiGRU generates future trajectories based on spatio-temporal embeddings. Results show that DAA-SGCN significantly improves prediction accuracy over baselines. Likewise, Jiang \etal~(2024)~\cite{jiang2024stmgf} proposed STMGF-Net, a spatio-temporal multi-graph fusion network for vessel trajectory forecasting in complex waters. The model constructs distinct graphs for vessel motion, risk, and attributes, fusing them using a spatio-temporal GCN and an attention-enhanced TCN for efficient feature extraction. STMGF-Net demonstrated significantly improved prediction accuracy compared to state-of-the-art graph-based methods.

While these graph-based methods were developed for multi-vessel prediction, they often suffer from model complexity and commonly rely on single-vessel metrics, such as the Average Displacement Error (ADE) and Final Displacement Error (FDE), which fail to capture inter-vessel dynamics during evaluation~\cite{weng2023joint}. Furthermore, these approaches often do not effectively capture vessel motion dynamics, neglecting physics-derived features, such as acceleration, jerk, and bearing, which are vital for modeling kinematic changes, particularly under external factors, such as strong winds or ocean currents~\cite{alam2022clustering, alam2024enhancing}.

Moreover, these approaches did not assess future collision risks based on predictions, despite this being a key goal of short-term trajectory forecasting. Perera \etal~\cite{perera2015collision} presented a distance-based simulation for two-vessel encounters using an extended Kalman filter to estimate the relative course, speed, and bearing. Ma \etal~\cite{ma2020data} framed risk prediction as a supervised time-series classification task, employing AIS-derived relative motion features and an attention-based BiLSTM for early risk detection. Tritsarolis \etal~\cite{tritsarolis2023collision} proposed a deep learning–based VCRA/F framework to estimate current risk and forecast future risk using real-world AIS data and a collision risk index. However, no established guidelines exist for the systematic identification of neighboring vessels in real-time and the simulation of their potential collision risks within a defined operational area.


While recent advances in deep learning, particularly graph-based and Transformer models, have improved vessel trajectory forecasting, important gaps remain. Most approaches focus on single-vessel prediction and evaluate performance with displacement errors that fail to capture inter-vessel dependencies. Graph-based methods address interactions but often suffer from high complexity, limited scalability, and the neglect of physics-derived features critical for modeling realistic vessel dynamics. Moreover, few works integrate trajectory forecasting with systematic collision risk assessment, leaving their practical value for navigational safety uncertain. Our work addresses these limitations by proposing a Transformer-based framework that jointly predicts the trajectories of multiple vessels, explicitly incorporates kinematic and physics-derived features, evaluates predictions with joint error metrics, and extends forecasting to collision risk analysis using DCPA and TCPA. This positions the framework as both methodologically novel and operationally relevant to enhance maritime safety.

\section{Preliminaries and Notation}
\label{sec:def-and-notation}

A \textit{\textbf{trajectory}}, \( \mathcal{T} = \{p_1, p_2, \dots, p_n \} \), is a sequence of points generated by a moving object, in which each point \( p_t \) represents a coordinate \( (x_t, y_t) \) and its kinematic state at time \( t \), where \( n \) denotes the number of points in \( \mathcal{T} \). 

In this study, a point \( p_t \) in a vessel trajectory is defined by the features listed in Table~\ref{tab:trajectory_symbols}.

\begin{table}[htbp]
\centering
\caption{Vessel Trajectory Notations}
\label{tab:trajectory_symbols}
    \begin{tabular}{@{}cl@{}}
        \toprule
        \textbf{Symbol} & \textbf{Description} \\
        \midrule
        \( x_t \) & Longitude (degrees) \\
        \( y_t \) & Latitude (degrees) \\
        \( v_t \) & Speed Over Ground (SOG, m/s) \\
        \( \psi_t \) & Course Over Ground (COG, degrees) \\
        \( a_t \) & Acceleration (m/s\(^2\)) \\
        \( \dot{\psi}_t \) & Rate of change of COG (degrees/s) \\
        \( j_t \) & Jerk (rate of change of acceleration, m/s\(^3\)) \\
        \( \beta_t \) & Bearing (vessel heading, degrees) \\
        \( \dot{\beta}_t \) & Bearing rate (rate of change of bearing, degrees/s) \\
        \bottomrule
    \end{tabular}
\end{table}
 
\noindent Accordingly, for each point \( p_t \), the input feature vector for the neural network model is defined as its transpose $p{_t}^T = \mathbf{x}_t$. 
\begin{equation}
    \mathbf{x}_t = [x_t, y_t, v_t, \psi_t, a_t, \dot{\psi}_t, j_t, \beta_t, \dot{\beta}_t]^T
\end{equation}

\noindent \textit{\textbf{Multi-Vessel Prediction}} involves forecasting future trajectories of a target vessel and its neighbors. For a target vessel \( M \), a set of $K$ neighbors \( S = \{S_1, S_2, \dots, S_k\} \), and prediction horizon \( H \), the model aims to simultaneously predict the future trajectories for the target, \( \mathcal{T}_{\text{pred}}^M = \{ \hat{p}_{t+1}^M, \dots, \hat{p}_{t+H}^M \} \), and its neighbors, \( \{\mathcal{T}_{\text{pred}}^{S_k}\}_{k=1}^K \), where \( \mathcal{T}_{\text{pred}}^{S_k} = \{ \hat{p}_{t+1}^{S_k}, \dots, \hat{p}_{t+H}^{S_k} \} \). Given the observed trajectories of the target, \( \mathcal{T}_{\text{obs}}^M = \{ p_1^M, \dots, p_t^M \} \), and neighbors, \( \{\mathcal{T}_{\text{obs}}^{S_k}\}_{k=1}^K \), where \( \mathcal{T}_{\text{obs}}^{S_k} = \{ p_1^{S_k}, \dots, p_t^{S_k} \} \). \\

\noindent The set of \textit{\textbf{Nearest Neighbors}} $S = \{S_1, S_2, \dots, S_K\}$ of a target vessel $M$ at time $t$ is determined using buffer radius $R_{\text{buf}}$. Let $D_{M,h}$ denote the total distance traveled by $M$ over a past duration of length $h$. The radius $R_{\text{buf}}$ around the current position $p_t^M$ of $M$ is defined as follows:
\begin{equation}
R_{\text{buf}} = 2 \cdot D_{M,h}
\end{equation}

Doubling the distance traveled by $M$ ensures that vessels approaching from opposite directions remain within the navigational scene. Accordingly, a vessel $V_k$ is included in set $S$ if $V_k \ne M$ and $\text{dist}(p_t^{V_k}, p_t^M) \le R_{\text{buf}}$, where $p_t^{V_k}$ is the current position of vessel $V_k$ at time $t$.

\section{Collision Avoidance Framework}
\label{sec:framework}

\begin{figure}[htbp]
    \centering
    \includegraphics[width=0.98\linewidth, keepaspectratio]{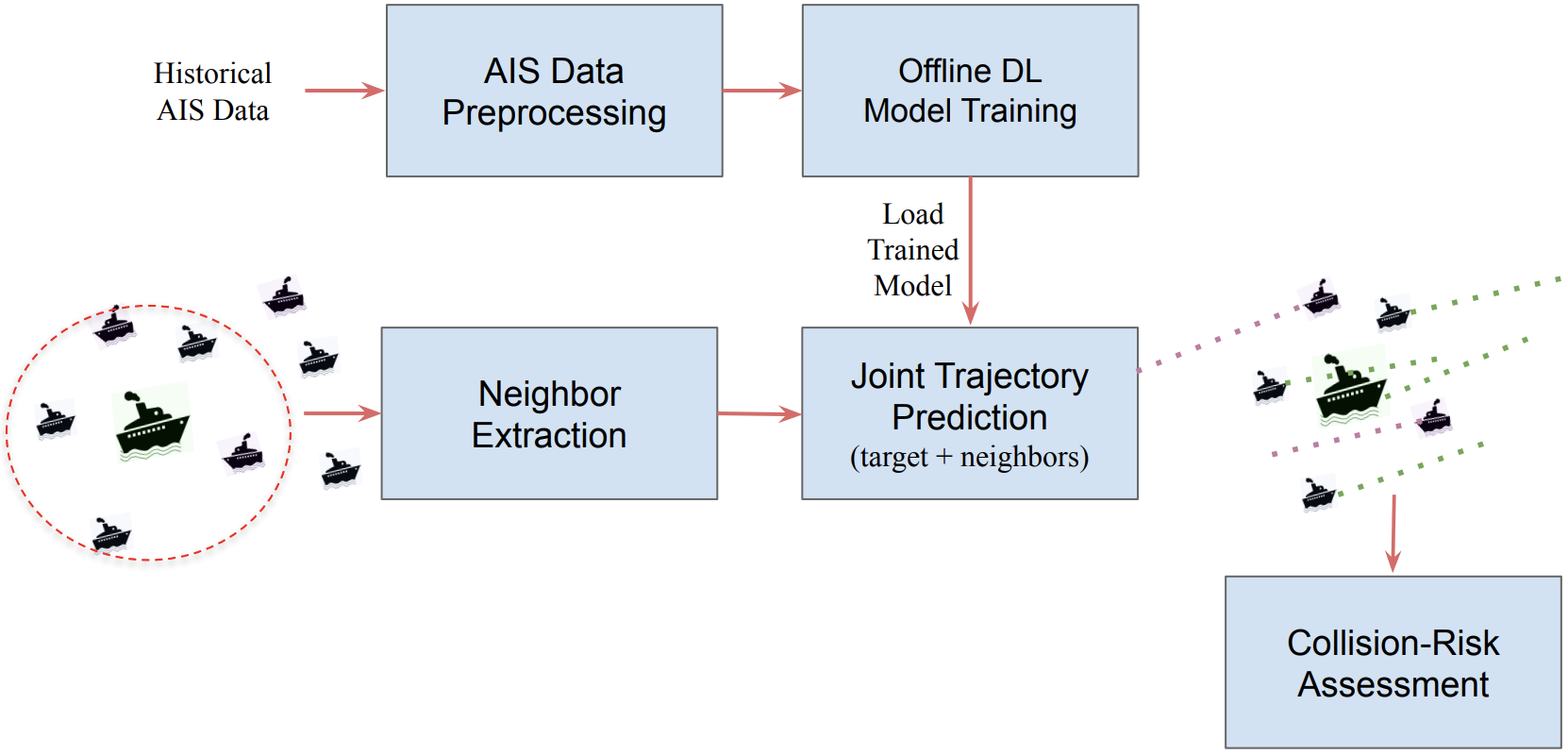}
    \caption{Overview of the proposed framework: identifying neighboring vessels, predicting their trajectories, and assessing collision risks.}
    \label{fig:pred-framework} 
\end{figure}

The pipeline of the prediction framework is shown in Fig.~\ref{fig:pred-framework}. We prepared a historical AIS dataset through a set of pre-processing steps to train
the proposed Transformer-based model offline. For prediction, the framework first identifies neighboring vessels from the future interaction area relative to the current position of the target vessel, and then predicts their future trajectories simultaneously using the trained model. As the final step of the process, future collision risks are assessed based on the predicted trajectories. 

\subsection{AIS Data Collection and Preprocessing}
\label{sub-sec:data-prep}

We collected raw historical AIS data from the Gulf of St. Lawrence, Canada (LON $-71.2$ to $-54.7$, LAT $44.9$ to $52.2$), shared by \textit{Spire.com} via AISViz/MERIDIAN~\cite{spadon2024maritime}, from January 2023 to June 2024. We extracted AIS messages with the following attributes: MMSI, timestamp, latitude, longitude, SOG, COG, and ship type. These messages were filtered by ship type and vessel identification number (MMSI), and the filtered messages were then grouped by MMSI to extract individual vessel trajectories. We prepared AIS data by applying the following steps to each extracted trajectory. \\

\subsubsection{Noise Filtering} Trajectories with invalid MMSI identifiers were excluded. AIS messages with duplicate timestamps or SOG below 0.5 knots (anchored or stationary vessels) were removed from the remaining trajectories. COG values were wrapped to the range $[0, 360)$ degrees. Only trajectories with at least 300 data points were retained. \\

\subsubsection{Trip Segmentation} Vessel trajectories contain AIS messages from multiple trips, including stationary periods in ports or during voyages. We split these trajectories into separate trips using time-based segmentation with a 60-min threshold. \\

\subsubsection{Trajectory Interpolation} Each trajectory segment was interpolated at 2-min intervals using \textit{Cubic Hermite splines} to ensure smooth transitions without overshooting or oscillations, while preserving the shape of the trajectory, making it well suited for capturing realistic vessel maneuvers. \\

\subsubsection{Physical Feature Engineering} A vessel's speed (\(v_t\)) and course (\(\psi_t\)) can be influenced by external factors such as strong winds, ocean currents, and other weather conditions. Consequently, to accurately capture vessel motion dynamics, additional physical features are critical. For each interpolated segment, we derived these physical features---acceleration (\(a_t\)), jerk (\( j_t \)), COG-rate (\(\dot{\psi}_t\)), bearing (\( \beta_t \)), and bearing-rate (\( \dot{\beta}_t \))---as follows, where \(\Delta t = t_i - t_{i-1}\) and \(\Delta x = x_i - x_{i-1}\).
\begin{equation}
\begin{aligned}
& a_i = \frac{v_i - v_{i-1}}{\Delta t}; \quad
  j_i = \frac{a_i - a_{i-1}}{\Delta t}; \\
& \dot{\psi}_i = \frac{\psi_i - \psi_{i-1}}{\Delta t}; \\
& \beta_i = \operatorname{atan2}\!\Big(
      \sin \Delta x_i \cos y_i, \\
& \qquad\quad \cos y_{i-1} \sin y_i 
      - \sin y_{i-1} \cos y_i \cos \Delta x_i
    \Big); \\
& \dot{\beta}_i = \frac{\beta_i - \beta_{i-1}}{\Delta t}
\end{aligned}
\end{equation}

\subsection{Transformer-based Prediction Model}
\label{sub-sec:proposed-model}

\begin{figure*}[htbp]
	\centering
	\includegraphics[width=0.85\linewidth, keepaspectratio]{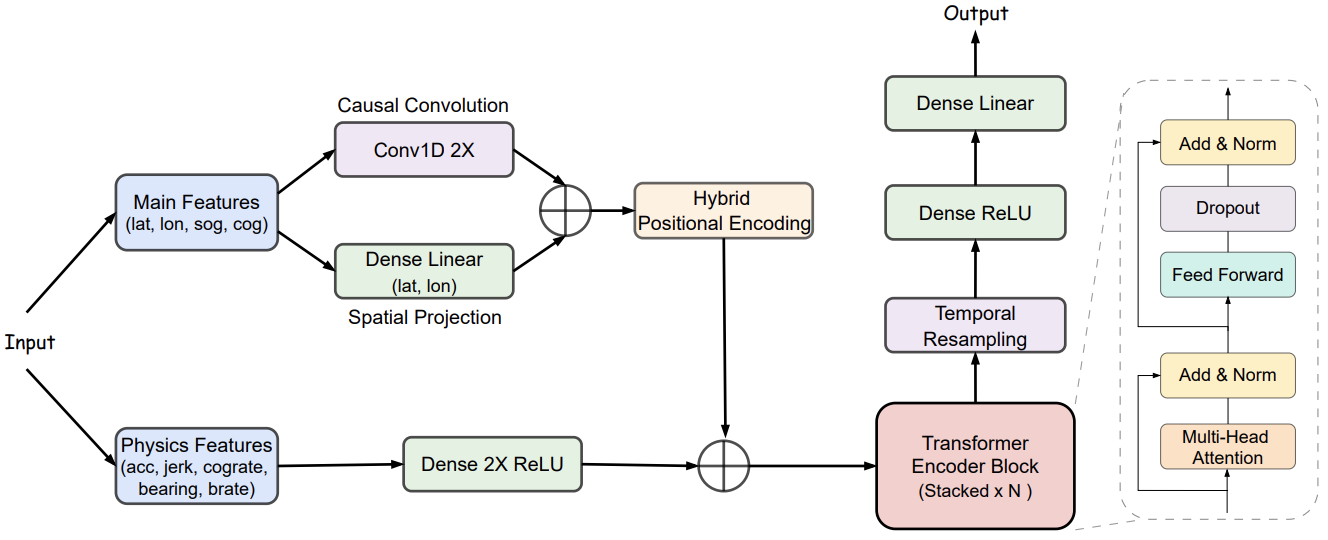}
	\caption{The proposed Transformer-based model for trajectory prediction.}
	\label{fig:proposed-transformer-model}
\end{figure*}

Figure~\ref{fig:proposed-transformer-model} illustrates the architecture of the proposed Transformer-based model. This model processes input data through two parallel streams: one handling core kinematic features (\({m}_t=[x_t, y_t, v_t, \psi_t]^T\)) using causal convolutions to capture the local temporal context, along with a spatial transformation to encode positional coordinates, and the other processes deriving physical features (\({p}_t=[a_t, \dot{\psi}_t, j_t, \beta_t, \dot{\beta}_t]^T\)) through non-linear transformations. The outputs from these spatially and temporally processed streams are fused via summation and enhanced using hybrid positional encoding to preserve the temporal order. The resulting combined sequence, which captures both spatial and temporal contexts, is then fed into a stack of Transformer encoder blocks to model the complex temporal dependencies and leverage encoded spatial relationships. Finally, a temporal resampling layer aligns the encoder output to the desired prediction time steps, followed by a series of transformations to generate the future trajectory. \\

\subsubsection*{Main Kinematic Features} 
The main navigation features, \({m}_t = [x_t, y_t, v_t, \psi_t]^T\), are passed through two parallel sub-modules, causal convolution and spatial projection. Causal \texttt{Conv1D} captures local temporal patterns within vessel trajectories, such as abrupt changes in course or speed, which are crucial for movement prediction. Causality is enforced by padding the input sequence to ensure that the convolutional kernel processes only past and current timestamps without access to future data. As latitude and longitude are angular and periodic, a spatial projection layer enables the model to learn a representation of geodetic coordinates, resulting in a more accurate understanding of spatial relationships. \\ 

\subsubsection*{Hybrid Positional Encoding} 
To capture the temporal order of observations within a trajectory, the model combines fixed sinusoidal~\cite{vaswani2017attention} and learned~\cite{irani2025positional} positional embeddings, enhancing its capacity to learn complex dependencies.

For a position \(pos\) in a trajectory and dimension \(i\) within the \(d_{\text{model}}\)-dimensional embedding space, sinusoidal encoding \(S_{\text{PE}}\) is defined as:
\begin{equation}
\begin{aligned}
    S_{\text{PE}}(pos, 2i)   &= \sin\!\left(\dfrac{pos}{10000^{\tfrac{2i}{d_{\text{model}}}}}\right), \\
    S_{\text{PE}}(pos, 2i+1) &= \cos\!\left(\dfrac{pos}{10000^{\tfrac{2i}{d_{\text{model}}}}}\right)
\end{aligned}
\end{equation}

Learned positional encoding \(L_{\text{PE}}\) is represented by an embedding layer \(E_{\text{pos}}\), which maps each position index to a \(d_{\text{model}}\)-dimensional vector:
\begin{equation}
    L_{\text{PE}}(pos) = E_{\text{pos}}(pos)
\end{equation}

During the model's forward pass, the hybrid encoding \(H_{\text{PE}}\) is computed by adding both encodings to the input vector \(\mathbf{x}_t\), which corresponds to the main kinematic features 
\( \mathbf{m}_t = [x_t, y_t, v_t, \psi_t]^T \):
\begin{equation}
    H_{\text{PE}} = \mathbf{m}_t + L_{\text{PE}} + S_{\text{PE}}
\end{equation}

\subsubsection*{Physics Features} 
The derived physical features, \({p}_t = [a_t, \dot{\psi}_t, j_t, \beta_t, \dot{\beta}_t]^T\), are processed through a non-linear dense layer to incorporate knowledge of vessel motion dynamics and physical constraints. These features serve as physical priors, enabling the model to learn trajectory evolution that adheres to motion laws. The resulting embedding is then fused with the positionally encoded navigation features and passed to the Transformer encoder for further modeling. \\

\subsubsection*{Transformer Encoder} 
This module~\cite{vaswani2017attention} takes as input a fused representation of physics embeddings and core navigational features, providing a comprehensive description of both the vessel’s current state and its dynamic motion trends. It captures long-range temporal dependencies and complex maneuvering patterns through multi-head self-attention and feed-forward networks. The output is a sequence of contextualized feature vectors that encode information ranging from low-level interactions to high-level motion patterns. \\

\subsubsection*{Temporal Resampling} This module adjusts the output sequence length from the encoder to align it with the prediction horizon. If the prediction horizon is longer than the input sequence length, \textit{upsampling} is performed; otherwise, \textit{downsampling}.
For upsampling, we used a 1D transposed convolution to increase the output sequence length. Alternatively, we can also use 1D upsampling to extend the sequence by repeating or interpolating features along the temporal dimension. The latter approach is often followed by a 1D convolution for refinement, ensuring smoothness and consistency in the upsampled output.
On the other hand, for downsampling, we can use either 1D convolution with a stride equal to the downsampling factor or 1D average pooling, which reduces the sequence length by averaging features over a temporal window. \\

\subsubsection*{Output Layer} Corresponds to a non-linear dense layer followed by a linear dense layer, which maps the resampled output to the predicted future trajectory coordinates, where the first layer introduces non-linearity to ensure robust feature space alignment to the output, and the linear layer directly yields latitude and longitude predictions. Therefore, the model inference is faster than standard Seq2Seq models.

\subsection{Trajectory Prediction and Collision Simulation}
\label{sub-sec:pred-collision}

As defined in Section~\ref{sec:def-and-notation}, to predict the future trajectories of a target vessel and its neighbors, we first compute a buffer area that represents the navigational scene around the target vessel’s current position and timestamp. The buffer radius is determined using a distance threshold based on the total distance traveled by the target vessel over the past duration---equal to the prediction horizon. To account for vessels approaching from the opposite direction, we doubled the buffer size (i.e., 2 $\times$ distance traveled). All vessels located within this buffer area at the current timestamp of the target vessel are considered neighbors of one another.
\begin{figure}[htbp]
    \centering
    \includegraphics[width=.98\linewidth]{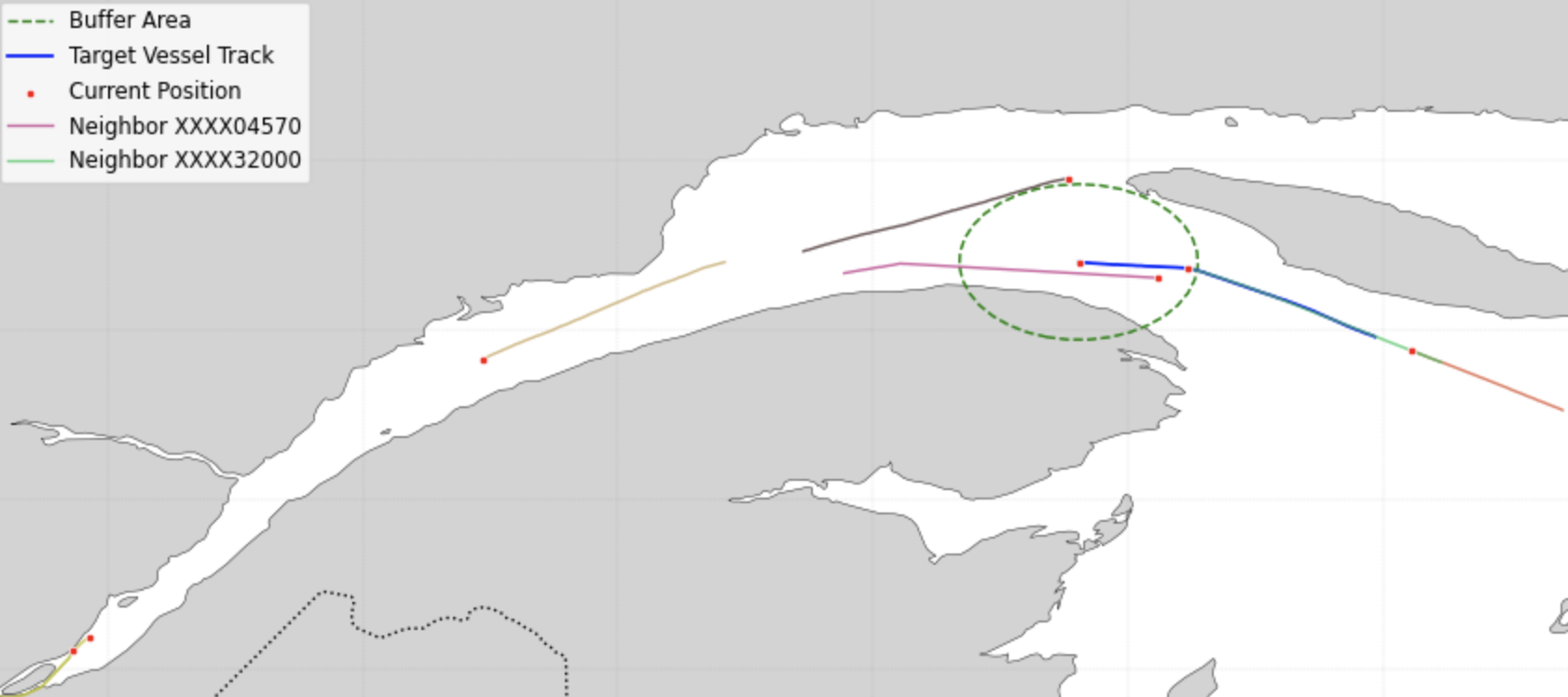}
    \caption{Multi-vessel navigation snapshot in the Gulf of St. Lawrence.}
    \label{fig:target-neighbors}
\end{figure}

Figure~\ref{fig:target-neighbors} illustrates a multi-vessel navigation scenario in the Gulf of St. Lawrence. It captures the relative positions of all vessels with respect to the current timestamp of the target vessel. This snapshot represents a typical input to our framework and is applicable at any given time. A buffer area, centered on the current position of the target vessel, defines the region of interest for analyzing vessel interactions. In this case, the buffer radius was computed as twice the distance traveled by the target vessel in the past $1$-hr. As shown, two neighboring ships were present within this area at that time.

After extracting neighbors, the framework loads the trained model and predicts the future trajectories of the target and neighboring vessels in parallel using DASK~\cite{rocklin2015dask}. In this study, the model uses the past $1$-hr trajectory of each vessel as input to predict $2$-hr in the future to analyze the joint prediction performance of the proposed model and assess potential collision risks in a multi-vessel scenario, which will be discussed in detail in the next section (Sections~\ref{sub_sec:joint-preds} and~\ref{sub_sec:collision}).

\section{Experimental Evaluation}
\label{sec:exp-evaluation}

\subsection{Experimental Setup}
We implemented this collision-avoidance framework in Python 3 using Keras/Tensorflow. \\

\subsubsection{AIS Dataset} 
\begin{figure}[htbp]
    \centering
    \includegraphics[width=0.8\linewidth]{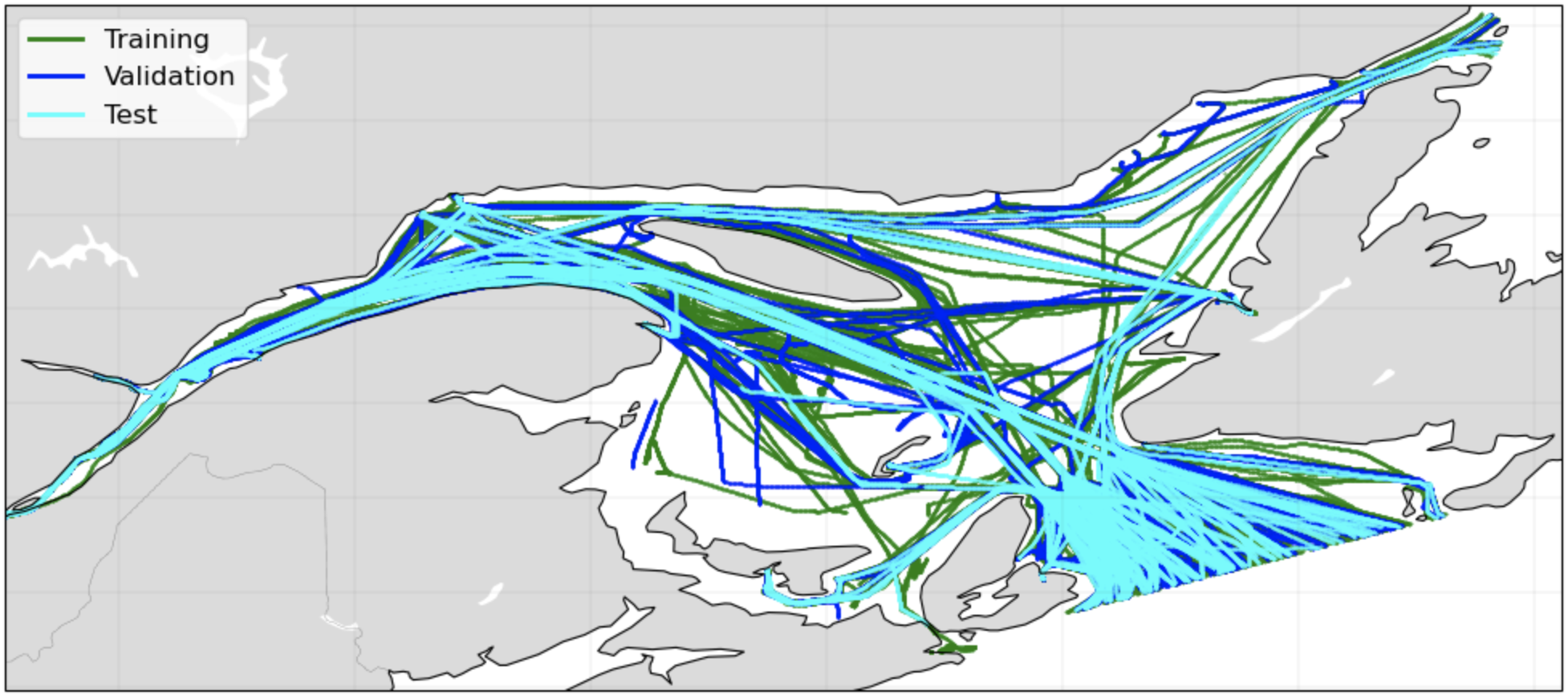}
    \caption{Spatial distribution of the tanker dataset in the Gulf of St. Lawrence. The test set comprises 10\% of the data, while the remaining 90\% is split into 80\% training and 20\% validation subsets.}
    \label{fig:tanker-dataset-split}
\end{figure}

We used tanker vessel trajectories in this evaluation. Following preprocessing, only trajectories with a minimum length of $5$ hrs were retained to accommodate our prediction window of $1$ hr of input for up to $3$ hrs of future prediction. The dataset consists of $2548$ tanker vessel trajectories, with a total of $2,507,491$ points. Figure~\ref{fig:tanker-dataset-split} shows the training, validation, and testing splits.

\subsubsection{Evaluation Metrics} 
To evaluate the efficacy of the prediction models, we used state-of-the-art single-agent metrics, namely, Average Displacement Error (ADE) and Final Displacement Error (FDE). However, as mentioned earlier, these metrics fail to assess the joint prediction performance in dynamic multi-vessel interacting scenarios. Consequently, we also used joint displacement metrics -- Joint ADE (JADE) and Joint FDE (JFDE)~\cite{weng2023joint}. To compute these metrics, we used the Haversine distance, \(D_{hv}\), as defined below, where \(R\) represents the Earth's radius, and \(\Delta x = x_2 - x_1\), \(\Delta y = y_2 - y_1\).
\begin{equation}
\small
D_{hv} = 2R \arcsin\! \sqrt{\, 
    \sin^{2}\!\Big(\tfrac{\Delta y}{2}\Big) 
    + \cos(y_{1}) \cos(y_{2}) \sin^{2}\!\Big(\tfrac{\Delta x}{2}\Big) 
}
\end{equation}

As defined in Eq.~\ref{eq:dist-metrics}, ADE represents the average distance between predicted and actual positions across all time steps, providing a comprehensive measure of trajectory accuracy. FDE, in contrast, measures only the distance between the final predicted and actual positions. Here, \(N\) is the total number of vessels, \(H\) is the prediction horizon, and \(K\) is the number of predicted trajectory samples. Existing approaches typically use \(min\) displacement errors to select the best prediction among \(K\) sampled trajectories. However, for a consistent and effective evaluation, we compute \(mean\) across all predicted samples. 
\begin{equation}
    \label{eq:dist-metrics}
    \small
    \begin{aligned}
        \text{ADE} &= \frac{1}{HN}  \sum_{v=1}^{N} \mean_{k=1}^{K} \sum_{t=1}^{H} D_{hv}\big((x_{t,v}^{(k)}, y_{t,v}^{(k)}), (\hat{x}_{t,v}, \hat{y}_{t,v})\big) \\ 
        \text{FDE} &= \frac{1}{N}  \sum_{v=1}^{N} \mean_{k=1}^{K} D_{hv}\big((x_{H,v}^{(k)}, y_{H,v}^{(k)}), (\hat{x}_{H,v}, \hat{y}_{H,v})\big)
    \end{aligned}
\end{equation}

In the case of a multi-vessel scenario, assessing the joint prediction performance of multiple interacting vessels is critical for collision-free navigation. Therefore, we used JADE and JFDE, as defined in~\ref{eq:jade_jfde}.  
\begin{equation}
    \label{eq:jade_jfde}
    \small
    \begin{aligned}
        \text{JADE} &= \frac{1}{HN} \mean_{k=1}^{K} \sum_{v=1}^{N} \sum_{t=1}^{H} D_{hv}\big((x_{t,v}^{(k)}, y_{t,v}^{(k)}), (\hat{x}_{t,v}, \hat{y}_{t,v})\big) \\
        \text{JFDE} &= \frac{1}{N} \mean_{k=1}^{K} \sum_{v=1}^{N} D_{hv}\big((x_{H,v}^{(k)}, y_{H,v}^{(k)}), (\hat{x}_{T,v}, \hat{y}_{T,v})\big)
    \end{aligned}
\end{equation}

While JADE and JFDE effectively measure joint prediction accuracy, they do not explicitly capture collision risk, particularly in congested waters, ports, or areas prone to accidents. It is equally important to ensure that the prediction models do not produce trajectories that lead to unsafe navigational situations. To address this, we used the \textit{Closest Point of Approach (CPA)}~\cite{zheng2022comprehensive} to simulate and analyze potential collision risks based on joint predictions. 

CPA represents the future point at which two vessels will be closest to each other. We compute the \textit{Distance at CPA (DCPA)} and compare it with a predefined safe distance, \(D_{\text{safe}}\). If \(\text{DCPA} \leq D_{\text{safe}}\), the encounter is flagged as a potential collision risk. Additionally, we compute the \textit{Time to CPA (TCPA)}, which indicates how long it will take to reach that closest point. This provides navigators with the opportunity to take proactive measures to avoid such collisions. \\

\subsubsection{Baselines} 
We selected several recently proposed DL-based approaches as baselines, summarized below (details in Section~\ref{sec:related-work}).

\begin{itemize}
    \item LSTM-Attn (2021)~\cite{capobianco2021deep}: A BiLSTM encoder–decoder with pooling and attention. 
    \item LSTM-FCL (2023)~\cite{chondrodima2023efficient}: An LSTM framework with trajectory augmentation.
    \item ConvLSTM (2023)~\cite{wu2023ship}: A ConvLSTM-based Seq2Seq model with spatial grid clustering.
    \item Multi-Gated-Attn (2024)~\cite{yang2024vessel}: An LSTM encoder with a multi-gated GRU attention decoder.
    \item Transformer (2017)~\cite{vaswani2017attention}: The original Transformer adapted for trajectory prediction. \\
\end{itemize}

\subsubsection{Model I/O} 
To prepare the datasets for training and testing the prediction models, we first applied min-max normalization to scale the dataset, followed by the sliding-window technique to extract input \(X\) and corresponding output \(Y\) sequences from a window of length \(w_{\text{in}} + w_{\text{out}}\). If \(t\) is the last time step of the input window, the input sequence is:
\begin{equation}
    X_t = \{\mathbf{x}_{t - w_{\text{in}} + 1}, \mathbf{x}_{t - w_{\text{in}} + 2}, \dots, \mathbf{x}_{t}\}
\end{equation}

The corresponding output sequence starts at time step \(t+1\), where \(w_{\text{out}}\) represents the prediction horizon (\(h\)):
\begin{equation}
    Y_t = \{\mathbf{y}_{t+1}, \mathbf{y}_{t+2}, \dots, \mathbf{y}_{t + w_{\text{out}}}\}
\end{equation} 

\subsubsection{Hyperparameter Settings} 
The proposed and baseline models were trained using the Adam optimizer for up to $100$ epochs, with a batch size of $64$. The models were optimized for Mean Squared Error (MSE), while performance was monitored using both MSE and Mean Absolute Error (MAE). Training was regularized with Early Stopping and the ReduceLROnPlateau scheduler based on validation performance. Furthermore, ModelCheckpoint was used to save the best model weights. While parameters for baseline models were tuned based on the information provided in the respective papers, those for our proposed Transformer-based model are as follows.

The 1D convolution block consists of $2$ layers with $128$ filters, a kernel size of $3$, and ReLU activation. The Transformer encoder comprises $4$ stacked layers with $128$ embedding dimensions, $8$ attention heads, $256$ feed-forward dimensions, and a dropout of $0.1$.

\subsection{Transformer-based Model Performance Analysis}

Table~\ref{tab:ade-fde-performance} reports the average ADE and FDE of $248$ tanker vessels for prediction horizons of up to $3$ hrs. In all cases, our proposed model consistently outperformed the recently presented models. For the $1$ hr horizon, ADE of some of these baselines, particularly LSTM-Attn~\cite{capobianco2021deep} and ConvLSTM~\cite{wu2023ship} Seq2Seq models, are relatively close to our proposed model. However, as the prediction horizon increases, our method demonstrates significantly lower displacement errors compared to all baselines, highlighting its robustness for long-term prediction as well.

\begin{table}[htpb]
  \centering
  \captionsetup{justification=centering}
  \caption{Displacement errors (meter) for different prediction horizons (Dataset: Tanker Vessels, The Gulf of St. Lawrence)}
  \label{tab:ade-fde-performance}
  \resizebox{\linewidth}{!}{%
  \begin{tabular}{lcccccc}
    \toprule
    \multirow{2}{*}{Model} 
      & \multicolumn{2}{c}{1\,hr} 
      & \multicolumn{2}{c}{2\,hr} 
      & \multicolumn{2}{c}{3\,hr} \\
    \cmidrule(lr){2-3} \cmidrule(lr){4-5} \cmidrule(lr){6-7}
      & ADE & FDE & ADE & FDE & ADE & FDE \\
    \midrule
    LSTM-Attn (2021)\cite{capobianco2021deep} & 1321 & 2684 & 2968 & 6667 & 5106 & 12115  \\
    LSTM-FCL (2023)\cite{chondrodima2023efficient} & 2402 & 4028 & 4141 & 7623 & 5948 & 11367  \\
    ConvLSTM (2023)\cite{wu2023ship} & 1305 & 2752 & 2991 & 6646 & 5048 & 11669  \\
    Multi-Gated-Attn (2024)\cite{yang2024vessel} & 8451 & 14847 & 14423 & 25383 & 19799 & 35372  \\
    Transformer (2017)\cite{vaswani2017attention} & 1365 & 2420 & 2734 & 5725 & 4419 & 9735 \\ 
    \midrule
    Proposed        & \textbf{1147} & \textbf{1537} & \textbf{1702} & \textbf{3002} & \textbf{2464} & \textbf{5027}   \\
    \bottomrule
  \end{tabular}
  }
\end{table}

\begin{figure}[htbp]
    \centering
    \includegraphics[width=0.92\linewidth]{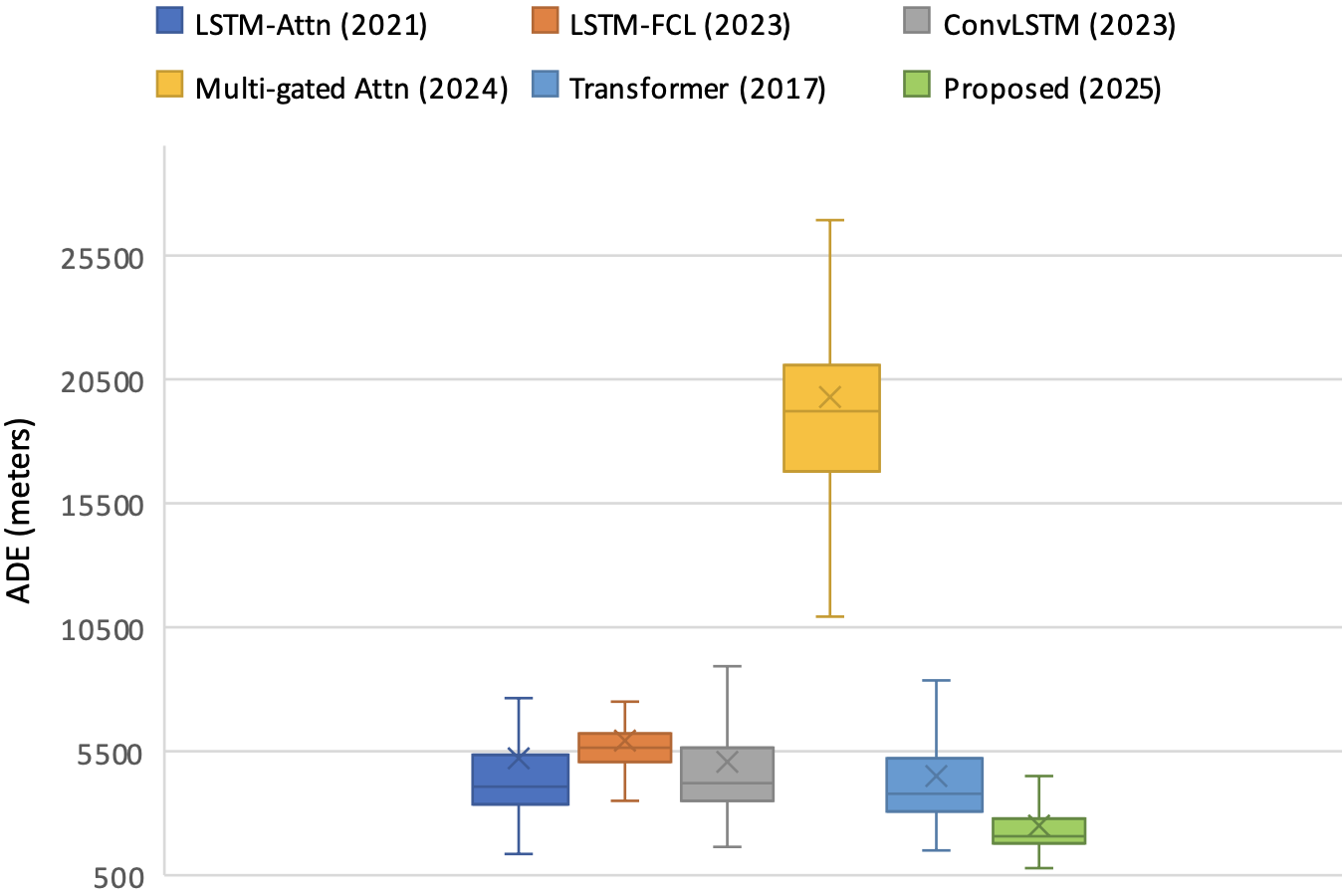}
    \caption{Box plot of ADE across DL models (\(W_{in} = 1\) hr, \(H = 3\) hr).}
    \label{fig:ADE}
\end{figure}

\begin{figure}[htbp]
    \centering
    \includegraphics[width=0.92\linewidth]{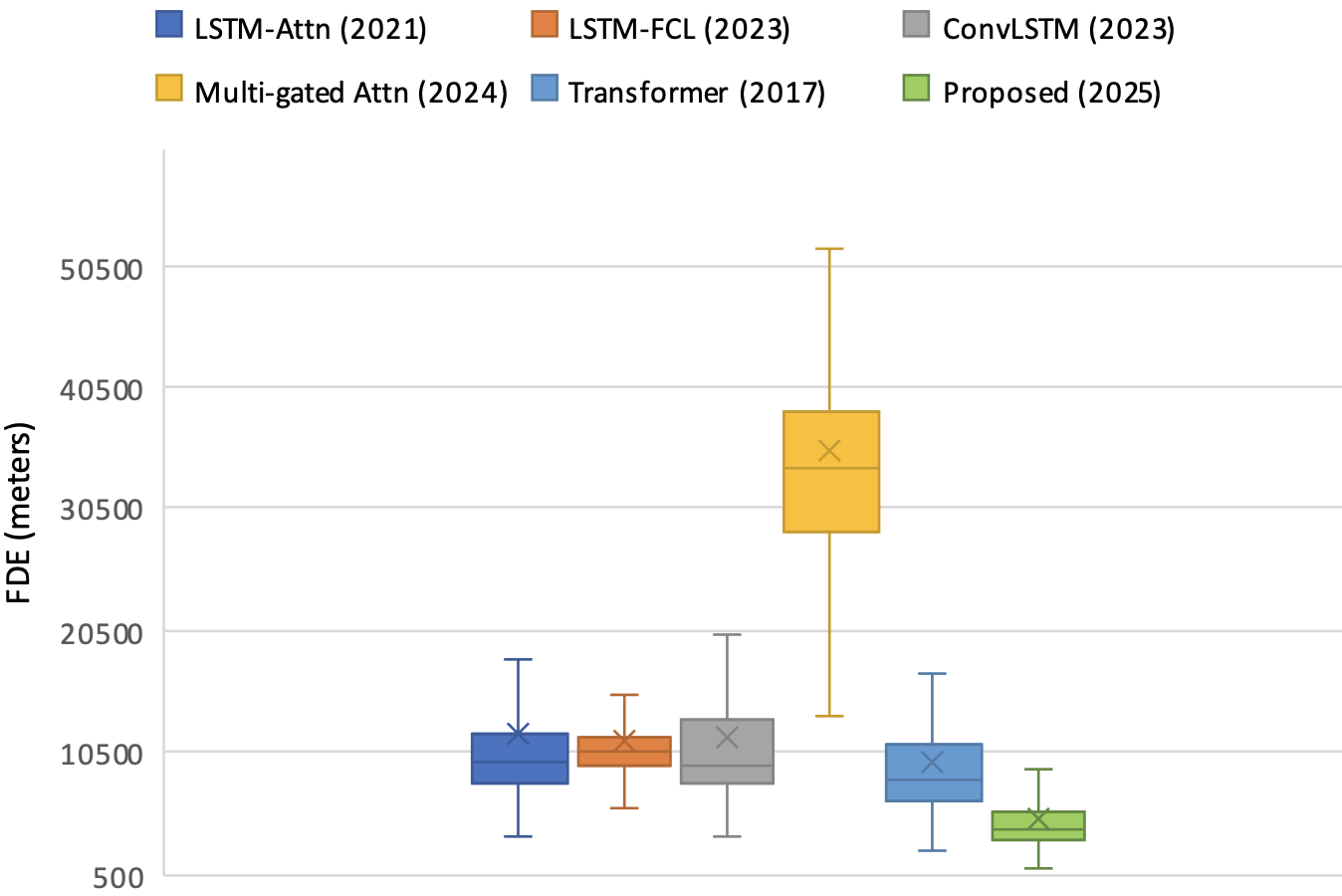}
    \caption{Box plot of FDE across DL models (\(W_{in} = 1\) hr, \(H = 3\) hr).}
    \label{fig:FDE}
\end{figure}

Figures~\ref{fig:ADE} and~\ref{fig:FDE} display box plots of ADE and FDE for a $3$hr prediction horizon, respectively. The proposed model consistently exhibited significantly lower mean and median errors than other methods. Its narrow IQR and short whiskers across both metrics not only indicate enhanced accuracy but also demonstrate greater consistency and robustness. Conversely, the multi-gated attention (2024)~\cite{yang2024vessel} method shows higher variability with wider dispersion and pronounced outliers, suggesting overfitting and poor generalization. Other methods demonstrated moderate performance but with greater variance than the proposed approach. \\

\noindent \textbf{Ablation Study:} To quantify the contribution of each module in our proposed architecture,  we conducted an ablation study, as summarized in Table~\ref{tab:ablation-study}. Removing the physics branch resulted in a performance drop (ADE: 2500, FDE: 5069), highlighting its subtle yet crucial role in maintaining physical consistency in predictions~\cite{Alam2025_PINN}. Eliminating both the physics and main branches while retaining hybrid positional encoding with all features (\(\mathbf{x}_t\)) as input further degrades performance (ADE: 2628, FDE: 5191), highlighting the importance of explicit feature modeling. When hybrid encoding was applied with only main features (\({m}_t\)) as input, the performance deteriorated significantly (ADE: 5519, FDE: 8213). Substituting hybrid encoding with sinusoidal encoding led to further degradation (ADE: 6240, FDE: 8738), thereby emphasizing the effectiveness of hybrid encoding. Compared to the vanilla Transformer (2017)~\cite{vaswani2017attention}, our model significantly outperforms in both ADE and FDE, demonstrating the value of our domain-specific architectural enhancements.

\begin{table}[htpb]
  \centering
  \captionsetup{justification=centering}
  \caption{Ablation Study ($W_{in}$ = 1{hr}, $H$ = 3{hr}, Unit = {Meter})}
  \label{tab:ablation-study}
  \begin{tabular}{lrr}
    \toprule
    \textbf{Model Variant} & \textbf{ADE} & \textbf{FDE} \\
    \midrule
    Proposed (full) & 2464 & 5027 \\
    \addlinespace
    
    \makecell[l]{Proposed (-physics branch, main features)} & 2500 & 5069 \\
    \addlinespace
    
    \makecell[l]{Proposed (-physics branch, -main branch,\\ +hybrid encoding, all features)} & 2628 & 5191 \\
    \addlinespace
    
    \makecell[l]{Proposed (-physics branch, -main branch,\\ +hybrid encoding, main features)} & 5519 & 8213 \\
    \addlinespace
    
    \makecell[l]{Proposed (-physics branch, -main branch,\\ +sinusoidal encoding, all features)} & 6240 & 8738 \\
    
    \bottomrule
  \end{tabular}
\end{table}

\subsection{Joint Prediction Performance Analysis}
\label{sub_sec:joint-preds}

\begin{figure}[!h]
    \centering
    \includegraphics[width=0.98\linewidth]{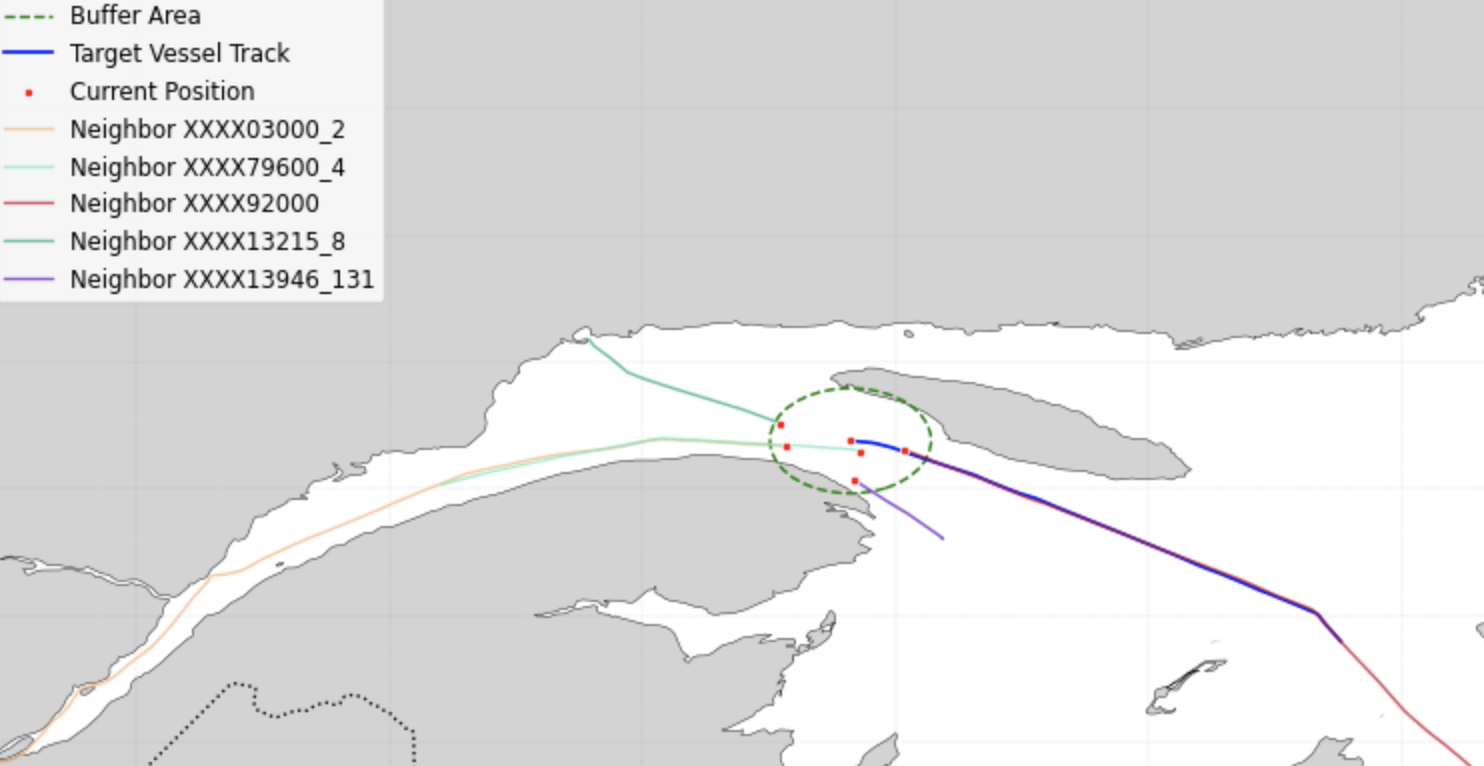}
    \caption{A state of a target and its neighbors within the defined buffer area.}
    \label{fig:joint-preds-1}
\end{figure}

To evaluate the efficacy of the proposed framework in a multi-vessel scenario, we conducted a joint prediction performance analysis using a representative real-world snapshot of a maritime environment, as depicted in Figure~\ref{fig:joint-preds-1}. This snapshot illustrates the current position and immediate past track of the target vessel, along with identified neighbors within a defined buffer area of interest.

\begin{figure}[!h]
    \centering
    \includegraphics[width=0.98\linewidth]{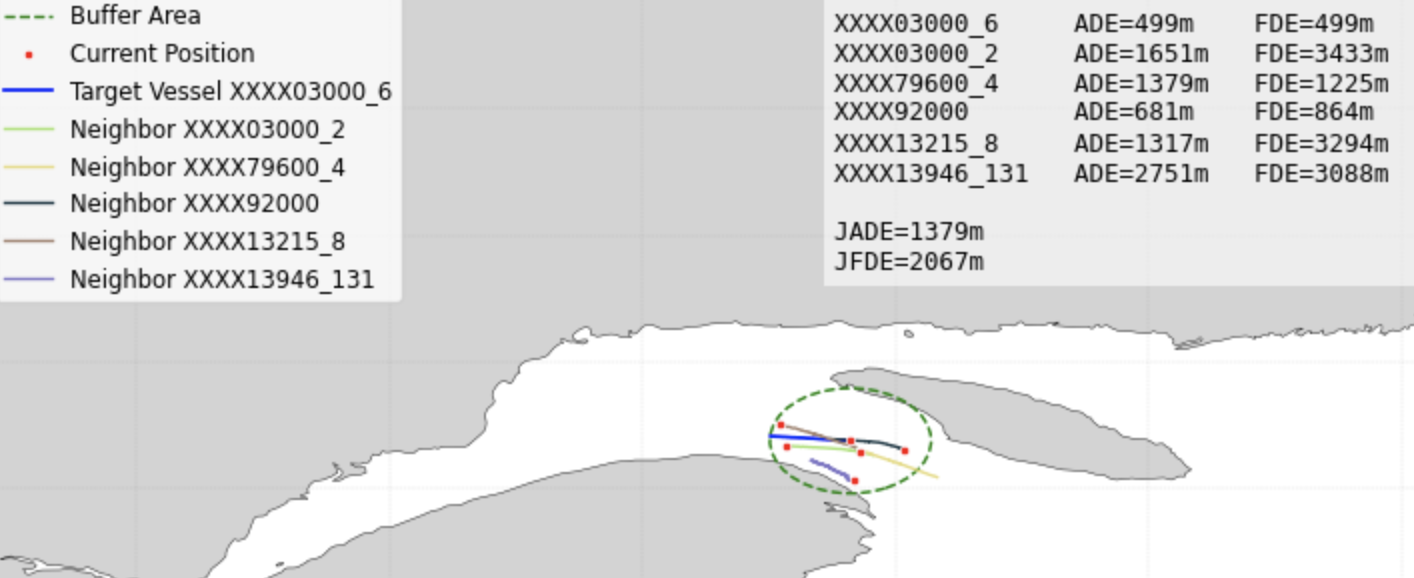}
    \caption{Joint predictions and the computed displacement errors.}
    \label{fig:joint-preds-2}
\end{figure}

Figure~\ref{fig:joint-preds-2} shows predicted trajectories over a $2$ hr horizon for each vessel, along with displacement errors (ADE and FDE). However, these metrics evaluate vessels individually and overlook inter-dependencies and temporal correlations in a multi-vessel setting, which can misrepresent actual prediction fidelity. For example, a model may achieve low ADE/FDE for a vessel (\textit{e.g.}, 681m/864m in this snapshot), yet its joint predictions could still lead to collisions or unrealistic relative movements among vessels.

To provide a more comprehensive assessment of prediction accuracy across multiple vessels in the area of interest, we computed joint displacement errors, JADE, and JFDE. These aggregate metrics explicitly consider inter-vessel interactions. JADE captures collective group behavior and interaction fidelity over the entire prediction horizon, whereas JFDE assesses end-state inconsistencies across vessels.

As shown in Figure~\ref{fig:joint-preds-2}, while the target vessel exhibits individual ADE/FDE values of 499m/499m, and the lowest ADE/FDE among the neighbors are 681m/864m, the overall JADE and JFDE for all vessels in this scenario are 1379m and 2067m, respectively. These higher aggregate error values highlight the importance of evaluating joint behavior, providing a more realistic measure of performance in complex, interactive maritime environments than individual metrics alone.

\subsection{Collision Risk Simulation and Analysis}
\label{sub_sec:collision}

\begin{figure}[htbp]
    \centering
    \includegraphics[width=0.98\linewidth]{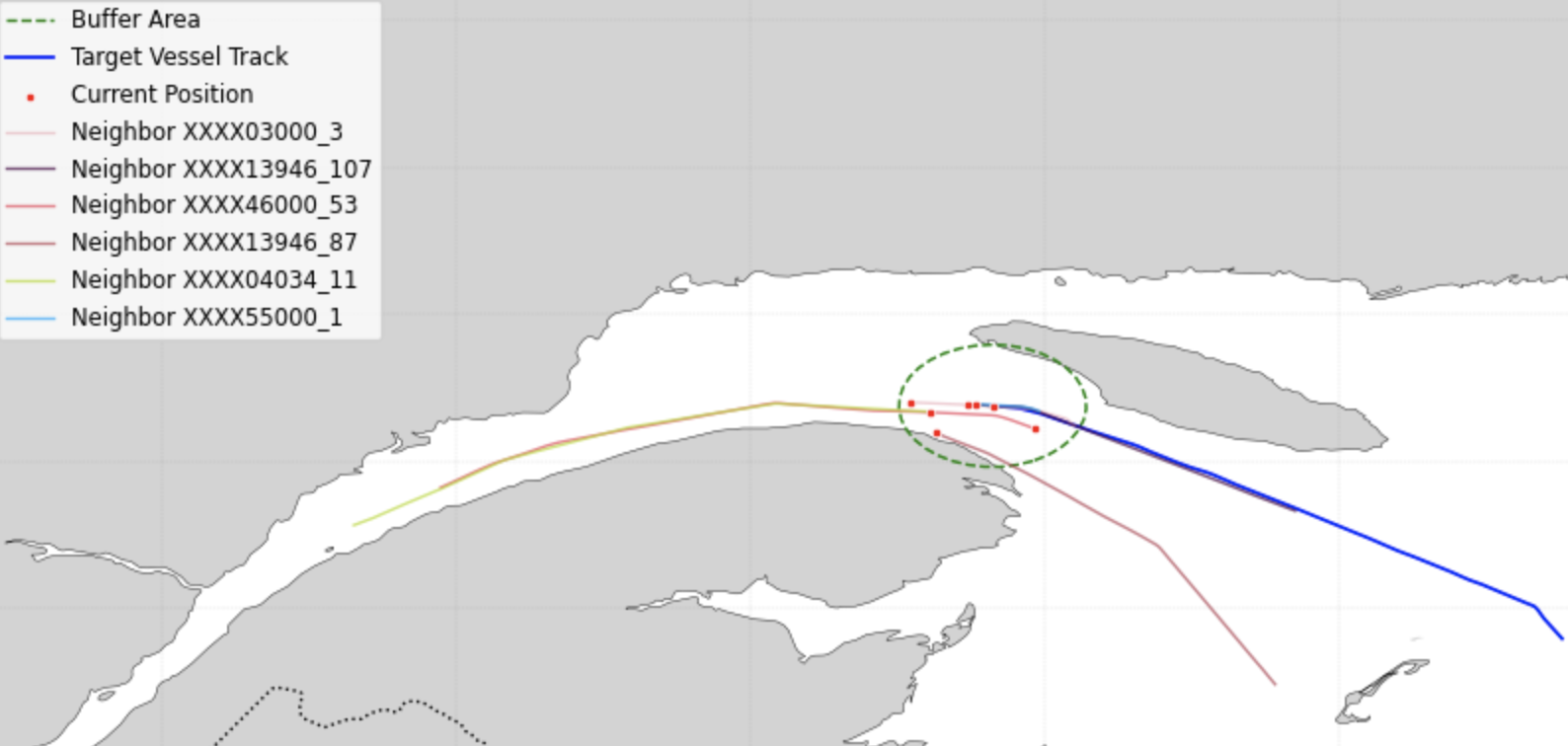}
    \caption{Navigation scenario -- neighbors within the buffer area.}
    \label{fig:collision-1}
\end{figure}

This section discusses the practical application of our prediction framework in real-world navigation scenarios. In particular, we demonstrate how it can assist navigators in identifying and mitigating potential collision risks based on the current state and immediate past trajectories of vessels in time and space. To such an end, Figure~\ref{fig:collision-1} illustrates a representative scenario, in which, for the current position and time of a target vessel, we first identify all neighboring vessels within a defined buffer area.

\begin{figure}[htbp]
    \centering
    \includegraphics[width=0.98\linewidth]{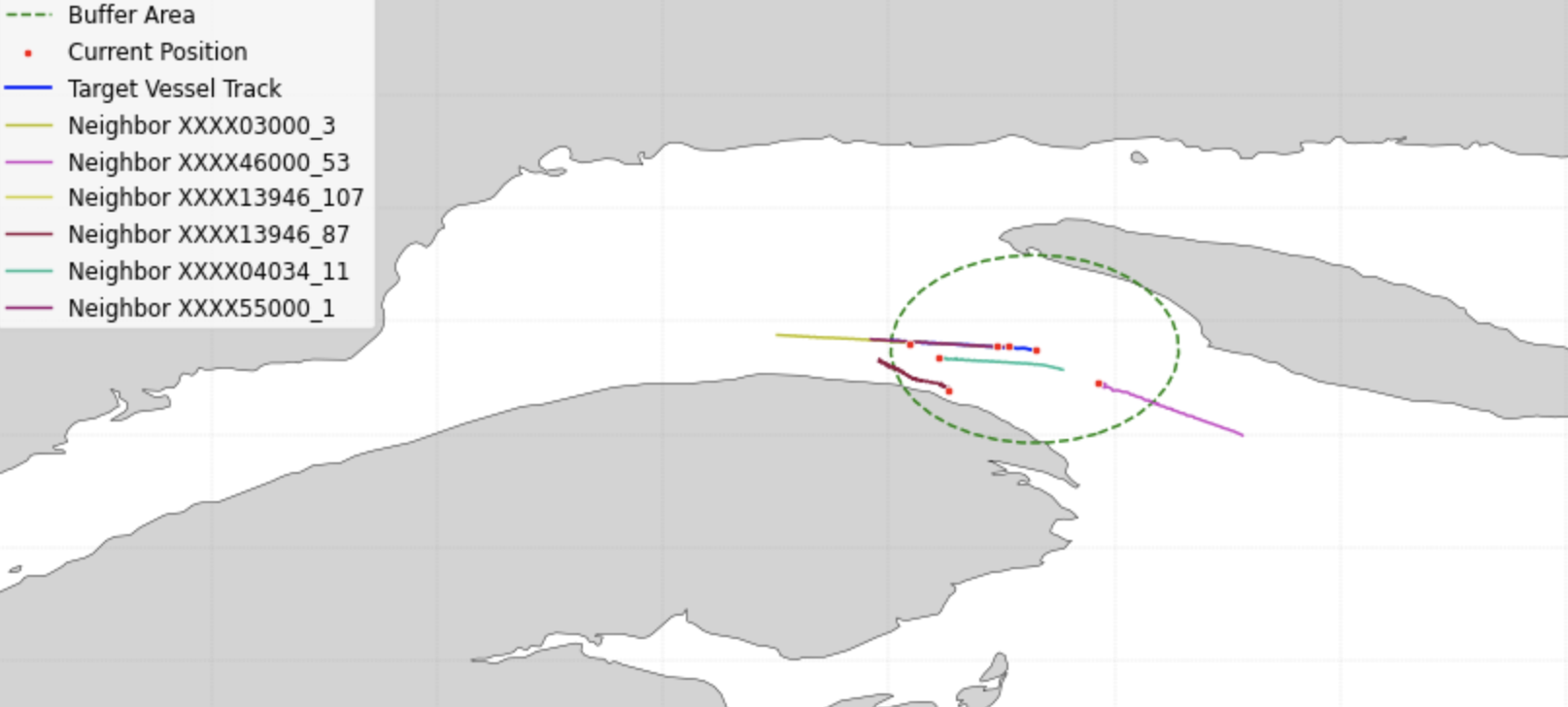}
    \caption{Predicted trajectories of the target vessel and its neighbors shown in Figure~\ref{fig:collision-1}}
    \label{fig:collision-2}
\end{figure}

To assess and warn about potential collision risks, our framework predicts the future trajectories of the target vessel and its neighbors in parallel, as shown in Figure~\ref{fig:collision-2}. Following trajectory prediction, we computed the DCPA for each target–neighbor vessel pair. DCPA is a critical metric in collision risk assessment, representing the estimated point at which the distance between two vessels reaches its minimum, assuming that they maintain their current course and speed. A potential collision risk is flagged if the calculated DCPA falls below a predefined safety threshold. In this study, we set the threshold at $500$ m, chosen based on the distances between successive points ($\approx [400,\text{m}, 600,\text{m}]$ over 2 min) and the average length of a vessel (a tanker, in this case).

\begin{figure}[htbp]
    \centering
    \includegraphics[width=0.98\linewidth]{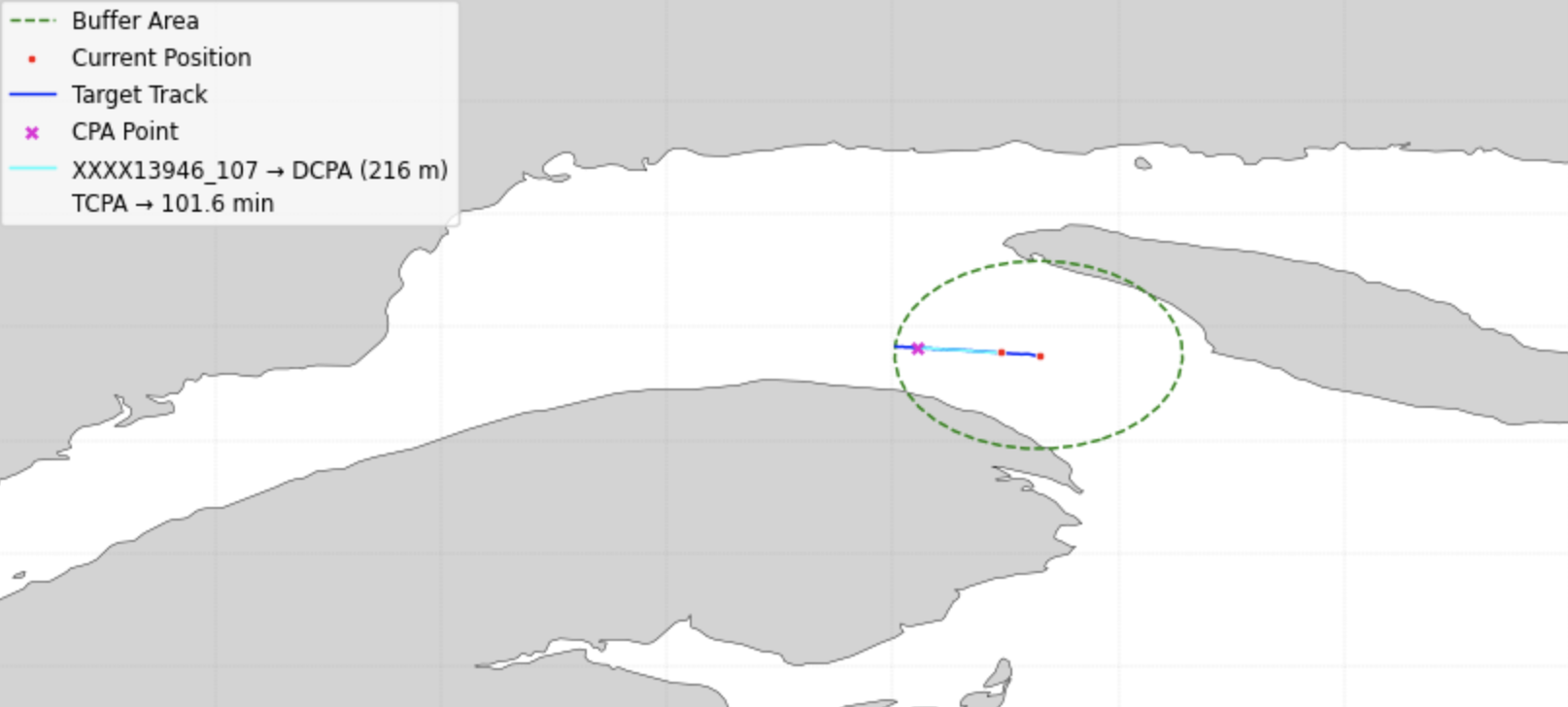}
    \caption{Simulation of a potential collision risk: vessel (MMSI XXXX13946) with DCPA $216m$  and TCPA $101.6$ min.}
    \label{fig:collision-3}
\end{figure}

As shown in Figure~\ref{fig:collision-3}, among the identified neighbors, one vessel (MMSI XXXX13946) is predicted to breach the safety threshold with the target vessel if both vessels maintain their current course and speed. The calculated DCPA for this pair is $226$ m, which is below the safety threshold. Concurrently, the TCPA is computed to estimate the time until the CPA occurs. For this specific vessel pair, the TCPA is $101.6$ min. Thus, navigators can use this information and warning to make proactive decisions to prevent potential collisions in the future.

\section{Conclusion}
\label{sec:conclusion}

This paper addressed the challenge of enhancing maritime situational awareness by jointly predicting multi-vessel trajectories and assessing collision risks. Unlike most prior works that focus on single-vessel forecasting, our framework explicitly models vessel interactions and integrates kinematic and physics-derived features into a Transformer-based architecture. By doing so, the model captures both short-term motion dynamics and long-range dependencies, producing physically consistent predictions.

Extensive experiments on over 2.5 million AIS records from tanker vessels in the Gulf of St. Lawrence demonstrated that the proposed approach significantly outperforms recent deep learning baselines, particularly at longer prediction horizons where displacement errors were reduced by up to 40\%. More importantly, by introducing joint evaluation metrics (JADE and JFDE) and extending predictions to collision risk analysis through DCPA and TCPA, the framework goes beyond accuracy benchmarks to address operational safety, offering a practical tool for decision support in real-world navigation.

Future research can build on several directions. First, extending the framework to heterogeneous vessel types (e.g., fishing, passenger, cargo) and diverse waterways would help assess generalization. Second, integrating external data such as weather, sea state, and port traffic could improve robustness, particularly in adverse scenarios. Third, collision risk assessment could be enhanced by moving beyond DCPA/TCPA to probabilistic risk models that account for prediction uncertainty. Fourth, model explainability remains an open challenge: developing interpretable attention mechanisms or physics-informed modules could improve trust in safety-critical deployments. Finally, deploying the framework in real-time platforms requires optimization for low-latency inference and distributed scalability, as well as human-in-the-loop studies.

\section*{Acknowledgments}
This research was partially supported by the \textit{National Council for Scientific and Technological Development} (CNPq 444325/2024-7), the Center for Artificial Intelligence (FAPESP 19/07665-4), the Natural Sciences and Engineering Research Council (NSERC RGPIN-2025-05179), and the Faculty of Computer Science at \textit{Dalhousie University} (DAL). The data used in this study was provided by AISViz/MERIDIAN and is subject to licensing restrictions, preventing the sharing of raw data. However, the pre-trained models and further code can be shared and are available upon request.

\balance
\bibliographystyle{IEEEtran}
\bibliography{IEEEabrv, IEEEreferences}

\vfill

\end{document}